\documentclass[runningheads]{llncs}

\usepackage{microtype}
\usepackage{amsfonts,amsmath,amssymb}
\usepackage{bbm}
\usepackage{booktabs}
\usepackage{color}
\usepackage{comment}
\usepackage{dsfont}
\usepackage{wrapfig}
\usepackage{makecell}
\usepackage{multirow}
\usepackage{tabularx}
\usepackage{multirow, booktabs}
\usepackage{siunitx}
\usepackage{xspace}
\usepackage{graphicx}
\usepackage{subcaption}

\let\originalparagraph\paragraph
\renewcommand{\paragraph}[2][.]{\originalparagraph{#2#1}}

\newcolumntype{L}{>{\raggedright\arraybackslash}X}
\newcolumntype{C}{>{\centering\arraybackslash}X}

\makeatletter
\DeclareRobustCommand\onedot{\futurelet\@let@token\@onedot}
\def\@onedot{\ifx\@let@token.\else.\null\fi\xspace}
\def\eg{\emph{e.g}\onedot} 
\def\ie{\emph{i.e}\onedot}

\def\etal{\emph{et al}\onedot}
\makeatother

\usepackage[width=122mm,left=12mm,paperwidth=146mm,height=193mm,top=12mm,paperheight=217mm]{geometry}

\newcommand{\mypara}[1]{\vspace{1.5mm}\noindent \textbf{#1}}
\newcommand{\titlecap}[2]{\textbf{#1} #2}

\begin{document}
\pagestyle{headings}
\mainmatter
\def\ECCVSubNumber{xxx}  

\title{Learning 3D Part Assembly from a Single Image} 


%
\author{Yichen Li\inst{*1} \and
Kaichun Mo\inst{*1} \and
Lin Shao\inst{1} \and
Minhyuk Sung\inst{2} \and
Leonidas Guibas\inst{1}}

%
\authorrunning{Li, et al.}
%
\institute{Stanford University \and Adobe Research}
\maketitle
\renewcommand{\thefootnote}{\fnsymbol{footnote}}\footnotetext[1]{:indicates equal contributions.}

\begin{abstract}
Autonomous assembly is a crucial capability for robots in many applications. For this task, several problems such as obstacle avoidance, motion planning, and actuator control have been extensively studied in robotics. However, when it comes to task specification, the space of possibilities remains underexplored. Towards this end, we introduce a novel problem, \emph{single-image-guided 3D part assembly}, along with a learning-based solution. We study this problem in the setting of \emph{furniture assembly} from a given complete set of parts and a single image depicting the entire assembled object.
Multiple challenges exist in this setting, including handling ambiguity among parts (\eg, slats in a chair back and leg stretchers) and 3D pose prediction for parts and part subassemblies, whether visible or occluded.
We address these issues by proposing a two-module pipeline that leverages strong 2D-3D correspondences and assembly-oriented graph message-passing to infer part relationships.
In experiments with a PartNet-based synthetic benchmark, we demonstrate the effectiveness of our framework as compared with three baseline approaches.


\end{abstract}

\section{Introduction}
\label{sec:intro}
\begin{figure}[t]
    \centering
    \vspace{-1mm}
    \includegraphics[width=0.9\linewidth]{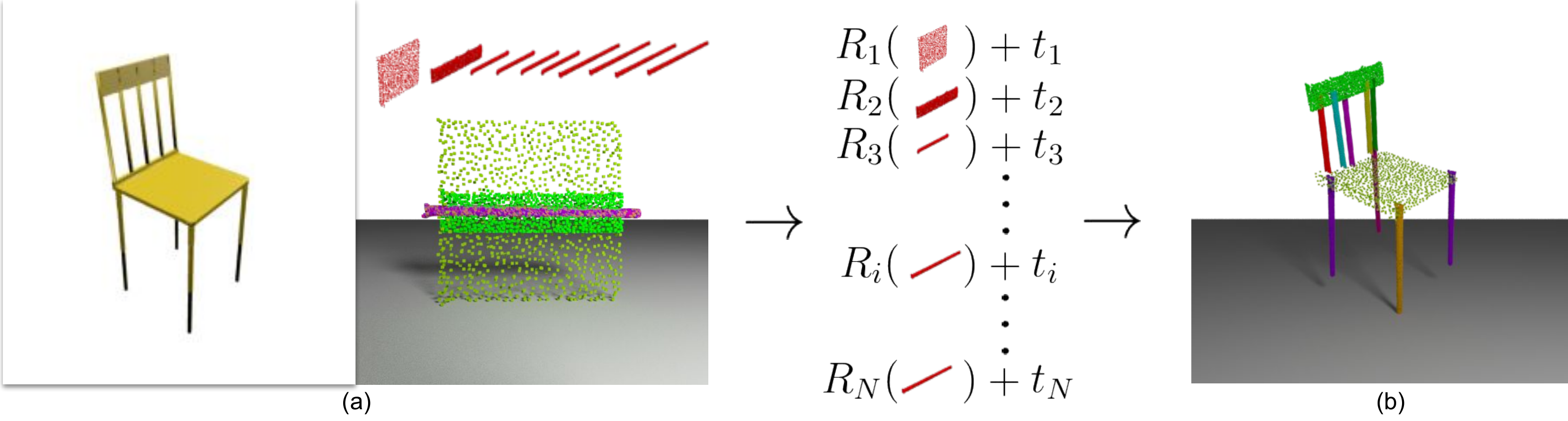}
    \vspace{-3mm}
    \caption{\titlecap{Single-Image-Based 3D Part Assembly Task.} Given as inputs an image and a set of part point clouds depicted in (a), the task is to predict 6D part poses in camera coordinates that assemble the parts to a 3D shape in the given image as shown in (b).}
    \label{fig:teaser}
    \vspace{-5mm}
\end{figure}

The important and seemingly straightforward task of furniture assembly presents serious difficulties for autonomous robots.
A general robotic assembly task consists of action sequences incorporating the following stages:
(1) picking up a particular part, 
(2) moving it to a desired 6D pose, 
(3) mating it precisely with the other parts, 
(4) returning the manipulator to a pose appropriate for the next pick-up movement.
Solving such a complicated high-dimensional motion planning problem~\cite{jimenez2013survey,hutchinson1990extending} requires considerable time and engineering effort. 
Current robotic assembly solutions first determine the desired 6D pose of parts~\cite{6225371} and then hard-code the motion trajectories for each specific object~\cite{Suarez-Ruizeaat6385}.
Such limited generalizability and painstaking process planning fail to meet demands for fast and flexible industrial manufacturing and household assembly tasks~\cite{levitin2006genetic}. 




To generate smooth and collision-free motion planning and control solutions, it is required to accurately predict 6D poses of parts in 3D space~\cite{Suarez-Ruizeaat6385,kaufman1996archimedes}. We propose a \emph{3D part assembly task} whose output can reduce the complexity of the high-dimensional motion planning problem. We aim to learn generalizable skills that allow robots to autonomously assemble unseen objects from parts~\cite{feng2015vision}. Instead of hand-crafting a fixed set of rules to assemble one specific chair, for example, we explore category-wise structural priors that helps robots to assemble all kinds of chairs. The shared part relationships across instances in a category not only suggest potential pose estimation solutions for unseen objects but also lead to possible generalization ability for robotic control policies~\cite{wangrobotic,lin2019learning,nelson1993visual,thorsley2004stereo}.

We introduce the task of \emph{single-image-guided 3D part assembly}: inducing 6D poses of the parts in 3D space~\cite{1013205} from a set of 3D parts and an image depicting the complete object.
Robots can acquire geometry information for each part using 3D sensing, but the only information provided for the entire object shape is the instruction image.
Different from many structure-aware shape modeling works~\cite{Mo:2019:Structurenet,Wu:2019:sagnet,Gao:2019,Wang:2018,Wu:2019:pqnet,Li:2020,Schor:2019},
we do not assume any specific granularity or semantics of the input parts, since the given furniture parts may not belong to any known part semantics and some of the parts may be provided pre-assembled into bigger units.
We also step away from instruction manuals illustrating the step-by-step assembling process, as teaching machines to read sequential instructions depicted with natural languages and figures is still a hard problem.

At the core of the task lie several challenges.
First, some parts may have similar geometry. For example, distinguishing the geometric subtlety of chair leg bars, stretcher bars, and back bars is a difficult problem.
%
Second, 3D geometric reasoning is essential in finding a joint global solution, where every piece fits perfectly in the puzzle. Parts follow a more rigid relationship graph which determines a unique final solution that emerges from the interactions between the geometries of the parts. 
%
Third, the image grounds and selects one single solution from all possible part combinations that might all be valid for the generative task. Thus, the problem is at heart a reconstruction task where the final assembly needs to agree to the input image. 
%
Additionally, and different from object localization tasks, \emph{the 3D Part Assembly Task} must locate all input parts, not only posing the parts visible in the image, but also hallucinating poses for the invisible ones by leveraging learned data priors.
One can think of having multiple images to expose all parts to the robot, but this reduces the generalizability to real-world scenarios, and might not be easy to achieve.
Thus, we focus on solving the task of single-image and category-prior-guided pose prediction. 

In this paper, we introduce a learning-based method to tackle the proposed \emph{single-image-guided 3D part assembly} problem.
Given the input image and a set of 3D parts, we first focus on 2D structural guidance by predicting an part-instance image segmentation to serve as a 2D-3D grounding for the downstream pose prediction.
To enforce reasoning involving fine geometric subtleties, we have designed a context-aware 3D geometric feature to help the network reason about each part pose, conditioned on the existence of other parts, which might be of similar geometry. 
%
Building on the 2D structural guidance, we can generate a pose proposal for each visible part and leverage these predictions to help hallucinate poses for invisible parts as well. 
Specifically, we use a part graph network, based on edges to encode different relationships among parts, and design a two-phase message-passing mechanism to take part relationship constraints into consideration in the assembly.

To best of our knowledge, we are the first to assemble \emph{unlabeled} 3D parts with a \emph{single image} input.
We set up a testbed of the problem on the recently released PartNet~\cite{Mo:2019:PartNet} dataset. 
We pick three furniture categories with large shape variations that require part assembly: Chair, Table and Cabinet.
We compare our method with several baseline methods to demonstrate the effectiveness of our approach. We follow the PartNet official train-test splits and evaluate all model performances on the unseen test shapes.
Extensive ablation experiments also demonstrate the effectiveness and necessity of the proposed modules: 2D-mask-grounding component and the 3D-message-passing reasoning component.

In summary, our contributions are:
\begin{itemize}
\vspace{-2mm}
\item we formulate the task of \emph{single-image-guided 3D part assembly};
\item we propose a two-module method, consisting of a part-instance image segmentation network and an assembly-aware part graph convolution network;
\item we compare with three baseline methods and conduct ablation studies demonstrating the effectiveness of our proposed method. 
\vspace{-3mm}
\end{itemize}










\section{Related Work}
\label{sec:related}
We review previous works on 3D pose estimation, single-image 3D reconstruction, as well as part-based shape modeling, 
and discuss how they relate to our task.

\mypara{3D Pose Estimation.} 
%
%
Estimating the pose of objects or object parts is a long-standing problem with a rich literature. 
Early in 2001, Langley~\etal~\cite{yoon2003real} attempted to utilize visual sensors and neural networks to predict the pose for robotic assembly tasks. 
Andy~\etal~\cite{zeng2017multi} built an robotic system taking multi-view RGB-D images as the input and predicting 6D pose of objects for Amazon Picking Challenge.
Recently, Litvak~\etal~\cite{8794226} proposed a two-stage pose estimation procedure taking depth images as input. 
In the vision community, there is also a line of works studying instance-level object pose estimation for known instances~\cite{brachmann2014learning,rad2017bb8,tekin2018real,kehl2017ssd,xiang2017posecnn,tejani2017latent,brachmann2016uncertainty} and category-level pose estimation~\cite{gupta2015aligning,papon2015semantic,braun2016pose,wang2019normalized,chen2020learning} that can possibly deal with unseen objects from known categories.
There are also works on object re-localization from scenes~\cite{zeng20173dmatch,izadinia2017im2cad,wald2019rio}.
Different from these works, our task takes as inputs unseen parts without any semantic labels at the test time, and requires certain part relationships and constraints to be held in order to assemble a plausible and physically stable 3D shape.


\mypara{Single-Image 3D Reconstruction.}
There are previous works of reconstructing 3D shape from a single image with the representations of voxel grids~\cite{Choy:2016,Tatarchenko:2017,Wang:2018:ocnn,Richter:2018}, point clouds~\cite{Fan:2017,Lin:2018,Insafutdinov:2018}, meshes~\cite{Pixel2Mesh,Pixel2Mesh++}, parametric surfaces~\cite{Groueix:2018}, and implicit functions~\cite{Chen:2019,Mescheder:2019,Park:2019,Saito:2019,Xu:2019}.
While one can consider employing such 2D-to-3D lifting techniques as a prior step in our assembly process so that the given parts can be matched to the predicted 3D shape, it can misguide the assembly in multiple ways. For instance, the 3D prediction can be inaccurate, and even some small geometric differences can be crucial for part pose prediction. Also, the occluded area can be hallucinated in different ways. In our case, the set of parts that should compose the object is given, and thus the poses of occluded parts can be more precisely specified. Given these, we do not leverage 3D shape generation techniques and directly predict the part poses from the input 2D image.

\mypara{Part-Based Shape Modeling.}
3D shapes have compositional part structures.
Chaudhuri~\etal~\cite{Chaudhuri:2011}, Kalogerakis~\etal~\cite{Kalogerakis:2012} and Jaiswal~\etal~\cite{Jaiswal:2016} introduced frameworks learning probabilistic graphical models that describe pairwise relationships of parts.
Chaudhuri and Koltun~\cite{Chaudhuri:2010}, Sung~\etal~\cite{Sung:2017} and Sung~\etal~\cite{Sung:2018} predict the compatibility between a part and a \emph{partial} object for sequential shape synthesis by parts.
Dubrovina~\etal~\cite{Dubrovina:2019}, PAGENet~\cite{Li:2020} and CompoNet~\cite{Schor:2019} take the set of parts as the input and generates the shape of assembled parts.
Different from these works that usually assume known part semantics or a part database, our task takes a set of unseen parts during the test time and we do not assume any provided part semantic labels.

GRASS~\cite{Li:2017}, Im2Struct~\cite{Niu:2018} and StructureNet~\cite{Mo:2019:Structurenet} learns to generate box-abstracted shape hierarchical structures.
SAGNet~\cite{Wu:2019:sagnet} and SDM-Net~\cite{Gao:2019} learn the pairwise relationship among parts that are subsequently integrated into a latent representation of the global shape.
G2LGAN~\cite{Wang:2018} autoencodes the shape of an entire object with per-point part labels, and a subsequent network in the decoding refines the geometry of each part.
PQ-Net~\cite{Wu:2019:pqnet} represents a shape as a sequence of parts and generates each part at every step of the iterative decoding process.
All of these works are relevant but different from ours in that we obtain the final geometry of the object not by directly decoding the latent code into part geometry but by predicting the poses of the given parts and explicitly assembling them. 
There are also works studying partial-to-full shape matching~\cite{litany2012putting,litany2016non,domokos2015realigning}.
Unlike these works, we use a single image as the guidance, instead of a 3D model.

\section{Problem Formulation}\label{sec:prod}
\label{sec:problem}
\vspace{-2mm}
We define the task of \emph{single-image-guided 3D part assembly}:
given a single RGB image $I$
of size $m\times m$ depicting a 3D object $S$ and a set of $N$ 3D part point clouds $\mathcal{P} = \left\{p_1, p_2, \cdots, p_N\right\}$ ($\forall i, p_i\in\mathbb{R}^{d_{pc}\times3}$), we predict a set of part poses $\left\{(R_i, t_i)\mid R_i\in\mathbb{R}^{3\times3}, t_i\in\mathbb{R}^{3}, i=1,2,\cdots,N\right\}$ in $SE(3)$ space.
After applying the predicted rigid transformation to all the input parts $p_i$'s,
the union of them reconstructs the 3D object $S$. 
We predict output part poses $\left\{(R_i, t_i)\mid i=1,2,\cdots,N\right\}$ in the camera space, following previous works~\cite{fan2017point,wang2018pixel2mesh}. 
In our paper, we use Quaternion to represent rotation and use $q_i$ and $R_i$ interchangeably. 

We conduct a series of pose and scale normalization on the input part point clouds to ensure \textit{synthetic-to-real} generalizability.
We normalize each part point cloud pose $p_i \in \mathcal{P}$ to have a zero-mean center and use a local part coordinate system computed using PCA~\cite{PCA}.
To normalize the global scale of all training and testing data, we compute Axis-Aligned-Bounding-Boxes (AABB) for all the parts and normalize them so that the longest box diagonal across all $p_i$'s of a shape has a unit length 
while preserving their relative scales. 
We cluster the normalized part point clouds $p_i$'s into sets of geometrically equivalent part classes $\mathcal{C} = \left\{C_1, C_2, \cdots, C_K \right\}$, where $C_1 = \{p_i\}_{i=1}^{N_1}$, $C_2 = \{p_i\}_{i=N_1+1}^{N_1+N_2}$, etc.
For example, four legs of a chair are clustered together if their geometry is identical.
This process of grouping indiscernible parts is essential to resolve the ambiguity among them in our framework. 
$\mathcal{C}$ is a disjoint complete set such that $C_k\cap C_l = \phi$ for every $C_k, C_l\in\mathcal{C}, k\neq l$ and $\cup_{k=1}^K C_k = \mathcal{P}$.
We denote the representative point cloud $p_j$ for each class $C_j \in \mathcal{C}$.

\section{Method}
\label{sec:method}
\begin{figure}[t]
    \centering
    \vspace{-1mm}
    \includegraphics[width=\linewidth]{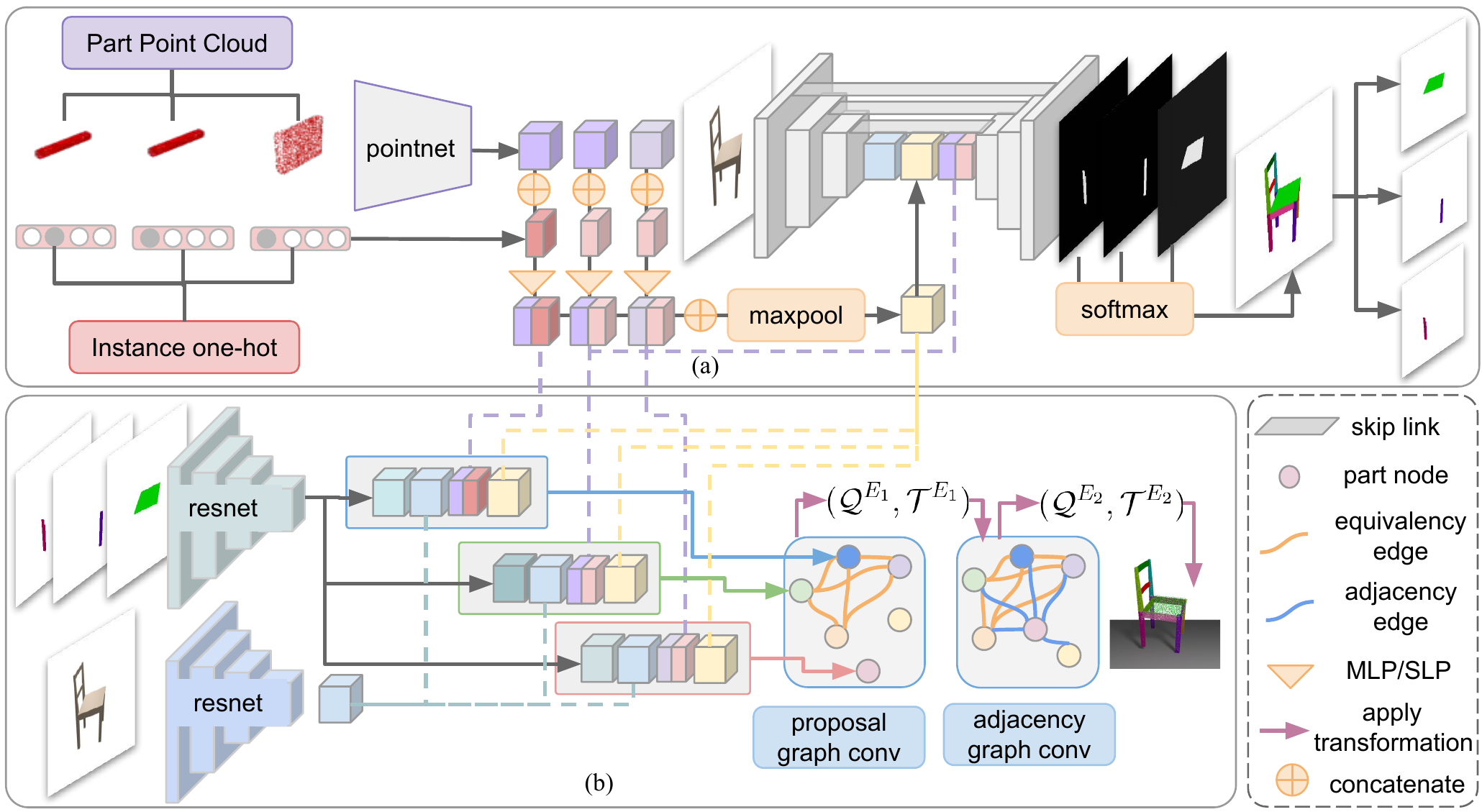}
    \vspace{-5mm}
    \caption{\titlecap{Network Architecture.} We propose a method that contains two network modules: (a) the part-instance image segmentation module, in which we predict 2D instance-level part masks on the image, and (b) the part pose prediction module, where we combine 2D mask information and 3D geometry feature for each part, push them through two phases of graph convolution, and finally predict 6D part poses.}
    \label{fig:arch}
    \vspace{-5mm}
\end{figure}

We propose a method for the task of \textit{single-image-guided 3D part assembly}, which is composed of two network modules: the part-instance image segmentation module and the part pose prediction module; see Figure~\ref{fig:arch} for the overall architecture.
We first extract a geometry feature of each part from the input point cloud $p_j \in \mathcal{C}$ and generates $N$ instance-level 2D segmentation masks $\left\{M_i\in\{0, 1\}^{m\times m}|i=1,2,\cdots,N\right\}$ on the input image ($m=224$). Conditioned on the predicted segmentation masks, our model then leverages both the 2D mask features and the 3D geometry features to propose 6D part poses $\left\{(q_i, t_i)|i=1,2,\cdots,N\right\}$.
We explain these two network modules in the following subsections. See supplementary for the implementation details. 
\vspace{-3mm}



\subsection{Part-Instance Image Segmentation}
To induce a faithful reconstruction of the object represented in the image, we need to learn a structural layout of the input parts from the 2D input. We predict a part instance mask $M_i\in\{0, 1\}^{m\times m}$ for each part $p_i$.
All part masks subject to the disjoint constraint, \ie, 
$M_{bg} + \sum_{i=1}^{N}M_i = \mathbf{1}$, where $M_{bg}$ denotes a background mask. 
If a part is invisible, we simply predict an empty mask and let the second network to halluciate a pose leveraging contextual information and learned data priors.
The task difficulties are two folds. First, the network needs to distinguish between the geometric subtlety of the input part point clouds to establish a valid 2D-3D correspondence. Second, for the identical parts within each geometrically equivalent class, we need to identify separate 2D mask regions to pinpoint their exact locations. 
Below, we explain how our proposed method is designed to tackle the above challenges. 

\mypara{Context-Aware 3D Part Features.}
To enable the network to reason the delicate differences between parts, we construct the context-aware 3D conditional feature $f_{3d}\in\mathbb{R}^{2F_2}$ ($F_2=256$), which is computed from three components: part geometry feature $f_{geo}\in\mathbb{R}^{F_2}$, instance one-hot vector $f_{ins}\in\mathbb{R}^{P_{max}}$ ($P_\text{max}=20$), and a global part contextual feature $f_{global}\in\mathbb{R}^{F_2}$.
We use PointNet~\cite{Qi:2017} to extract a global geometry feature $f_{geo}$ for each part point cloud $p_i$. 
If a part $p_j$ has multiple instances $k_j > 1$ within a geometrically equivalent class $\mathcal{C}_j$ (\eg four chair legs), we introduce an additional instance one-hot vector $f_{ins}$ to tell them apart.
For part which has only one instance, we use an one-hot vector with the first element to be 1.
For contextual awareness, we extract a global feature $f_{global}$ over all the input part point clouds, to facilitate the network to distinguish between similar but not equivalent part geometries (\eg a short bar or a long bar). 
Precisely, we first compute $f_{geo}$ and $f_{ins}$ for every part, then compute $f_{local}=SLP_1([f_{geo}; f_{ins}])\in\mathbb{R}^{F_2}$ to obtain per-part local feature, where SLP is short for Single-Layer Perception.
We aggregate over all part local features via a max-pooling symmetric function to compute the global contextual feature $f_{global}=SLP_2\left(MAX_{i=1,2,\cdots,N}\left(f_{i, local}\right)\right)$.
Finally, we define $f_{3d}=[f_{local}; f_{global}]\in\mathbb{R}^{2F_2}$ to be the context-aware 3D per-part feature.

\mypara{Conditional U-Net Segmentation.}
We use a conditional U-Net~\cite{unet} for the part-instance segmentation task.
Preserving the standard U-Net CNN architecture,
our encoder takes an 3-channel RGB image as input and produce a bottleneck feature map $f_{2d}\in\mathbb{R}^{F_1\times7\times7}$ ($F_1=512$). Concatenating the image feature $f_{2d}$ with our context-aware 3D part conditional feature $f_{3d}$, we obtain $f_{2d+3d}=[f_{2d}, f_{3d}]\in\mathbb{R}^{(F_1+2F_2)\times7\times7}$, where we duplicate $f_{3d}$ along the spatial dimensions for $7\times7$ times.
The decoder takes the conditional bottleneck feature $f_{2d+3d}$ and decodes a part mask $M_i$ for evert input part $p_i$.
We keep skip links as introduced in the original U-Net paper between encoder and decoder layers.
To satisfy the non-overlapping constraint, we add a SoftMax layer across all predicted masks, augmented with a background mask $M_{bg} \in \{0,1\}^{(m \times m )}$.

\subsection{Part Pose Prediction}
With the 2D grounding masks produced by the part-instance image segmentation module, we predict a 6D part pose $(R_i, t_i)$ for every input part $p_i \in \mathcal{P}$ using the part pose prediction module. 
We predict a unit Quaternion vector $q_i$ that corresponds to a 3D rotation $R_i$ and a translation vector $t_i$ denoting the part center position in the camera space.

Different from object pose estimation, the task of part assembly requires a joint prediction of all part poses. Part pose predictions should not be independent with each other, as part poses follow a set of more rigid relationships, such as symmetry and parallelism. For a valid assembly, parts must be in contact with adjacent parts.
The rich part relationships restrict the solution space for each part pose. 
We leverage a two-phase graph convolutional neural network to address the joint communication of part poses for the task of part assembly.

\mypara{Mask-Conditioned Part Features.}
We consider three sources of features for each part: 2D image feature $f_{img}\in\mathbb{R}^{F_3}$, 2D mask feature $f_{mask}\in\mathbb{R}^{F_3}$ ($F_3=512$), context-aware 3D part feature $f_{3d}\in\mathbb{R}^{2F_2}$.
We use a ResNet-18~\cite{He:2015} pretrained on ImageNet~\cite{deng2009imagenet} to extract 2D image feature $f_{img}$.
We use a separate ResNet-18 that takes the 1-channel binary mask as input and extracts a 2D mask feature $f_{mask}$, where masks for invisible parts are predicted as empty.
Then, finally, we propagate the 3D context-aware part feature $f_{3d}$ introduced in the Sec. 4.1 that encodes 3D part geometry information along with its global context.

\mypara{Two-Phase Graph Convolution.}
We create a part graph $\mathcal{G}=(V, E)$, treating every part as a node and propose a two-phase of graph convolution to predict the pose of each part. We first describe how we construct the edges in each phase, and then introduce our assembly-oriented graph convolution operations.

During the first phase, we draw pairwise edges among all parts $p_i$ in every geometrically equivalent part classes $C_j$ and perform graph convolution over $\mathcal{G}^1=(V, E^1)$, where
\vspace{-1.3mm}
\begin{equation}
    E^1=\left\{(p_{i_1}, p_{i_2})|\forall p_{i_1}, p_{i_2} \in C_j, i_1\neq i_2, \forall C_j \in\mathcal{C}\right\}.
\end{equation}
Edges in $E^1$ allow message passing among geometrically identical parts that are likely to have certain spatial relationships or constraints (\eg four legs of a chair have two orthogonal reflection planes). 
After the first-phase graph convolution, each node $p_i$ has an updated node feature. 
The updated node feature is then decoded as an 6D pose $(R_i, t_i)$ for each part. 
The predicted part poses produce an initial assembled shape. 

We leverage a second phase of graph convolution to refine the predicted part poses.
Besides the edges in $E^1$, we draw a new set of edges $E^2$ by finding top-5 nearest neighbors for each part based upon the initial assembly and define $\mathcal{G}^2=\left(V, E^1\cup E^2\right)$.
The intuition here is that once we have an initial part assembly, we are able to connect the adjacent parts so that they learn to attach to each other with certain joint constraints. 

We implement the graph convolution as two iterations of message passing~\cite{xu2018powerful,wang2019dynamic,Mo:2019:Structurenet}.
Given a part graph $\mathcal{G}=\left(V, E\right)$ with initial node features $f^0$ and edge features $e^0$, each iteration of message passing starts from computing edge features
\vspace{-0.5mm}
\begin{equation}
    e^{t+1}_{{(p_{i_1}, p_{i_2})}} = SLP_{g}\left([f_{i_1}^t; f_{i_2}^t; e^t_{(p_{i_1}, p_{i_2})}]\right), t \in \{0,1\}.
\end{equation}
where we do not use $e^0$ during the first phase of graph convolution, and define $e^0_{{(p_{i_1}, p_{i_2})}}=0$ if $(p_{i_1}, p_{i_2})\in E^1$ and $e^0_{{(p_{i_1}, p_{i_2})}}=1$ if $(p_{i_1}, p_{i_2})\in E^2$ for the second phase.
Then, we perform average-pooling over all edge features that are connected to a node and obtain the updated node feature
\vspace{-1.3mm}
\begin{equation}
    f^{t+1}_i=\frac{1}{\left|\left\{u\mid (p_i, p_u)\in E\right\}\right|}\sum_{(p_i, p_u)\in E}e^{t+1}_{(p_i, p_u)}, t \in \{0,1\}.
\end{equation}
We define $f^{t+1}_i=f^t_i$ if there is no edge drawn from node $i$. 
We define the final node features to be $f_i=[f^0_i; f^1_i; f^2_i]$ for each phase of graph convolution.

Respectively, we denote the final node feature of first phase and second phase graph convolution to be $^1f_i$ and $^2f_i$ for a part $p_i$.




\mypara{Part Pose Decoding.}
After gathering the node features after conducting the two-phase graph convolution operations as $^1f_{i}$ and $^2f_{i}, i \in \{1, 2, \cdots, N\}$,
we use a Multiple-Layer Perception (MLP) to decode part poses at each phase. 
\vspace{-1mm}
\begin{equation}
    ^sq_{i}, ^st_{i} = MLP_{PoseDec}\left(^sf_{i}\right), s \in \{1,2\}, i \in \{1, 2, \cdots, N\}.
\end{equation}
To ensure the output of unit Quaternion prediction, we normalize the output vector length so that $\left\lVert^sq_{i}\right\rVert_2=1$.

\subsection{Training and Losses}
We first train the part-instance image segmentation module until its convergence and then train the part pose prediction module.
Empirically, we find that having a good mask prediction is necessary before training for the part poses.

\mypara{Loss for Part-Instance Image Segmentation.}
We adapt the negative \textit{soft-iou} loss from \cite{softiou:2017} to supervise the training of the part-instance image segmentation module. 
We perform Hungarian matching~\cite{Kuhn:1955} within each geometrically equivalent class to guarantee that the loss is invariant to the order of part poses in ground-truth and prediction. The loss is defined as
\begin{equation}
\mathcal{L}_{mask_i} = - \frac{\sum_{{u,v \in [m, m]}} \hat{M}_{i}^{(u,v)} \cdot M_{\mathcal{M}(i)}^{(u,v)}}{\sum_{u,v \in [m, m]} \left(\hat{M}_{\mathcal{M}(i)}^{(u,v)}+M_{i}^{(u,v)}-\hat{M}_{\mathcal{M}(i)}^{(u,v)} \cdot M_{i}^{(u,v)}\right)}.
\end{equation}
where $M_{i}\in\{0,1\}^{[m, m]}$ and $\hat{M}_{\mathcal{M}(i)}\in[0,1]^{[m, m]}$ denote the ground truth and the matched predicted mask.
$\mathcal{M}$ refers to the matching results that match ground-truth part indices to the predicted ones.
$[m, m]$ includes all 2D index $(u,v)$'s on a $224\times 224$ image plane. 

\mypara{Losses for Part Pose Prediction.}
For the pose prediction module, we design an order-invariant loss by conducting Hungarian matching within each geometry-equivalent classes $C_i \in \mathcal{C}$. 
Additionally, we observe that separating supervision loss for translation and rotation helps stabilize training. We use the following training loss for the pose prediction module. 
\vspace{-1.3mm}
\begin{equation}
\mathcal{L}_{pose} =  \sum_{i=1}^N (\lambda_1 \times \mathcal{L}_{T} + \lambda_2 \times  \mathcal{L}_{C} + \lambda_3 \times \mathcal{L}_{E} ) + \lambda_4 \times \mathcal{L}_{W}
\end{equation}

We use the $L_2$ Euclidean distance to measure the difference between the 3D translation prediction and ground truth translation for each part. We denote $\mathcal{M}$ as the matching results.
\begin{equation}
    \mathcal{L}_{T_i} = \|\hat{t}_{\mathcal{M}(i)}- t_i\|_2, \forall i \in  \{1, 2, \cdots, N\}.
\end{equation}
where $\hat{t}_{\mathcal{M}(i)}$ and $t_i$ denote the matched predicted translation and the ground truth 3D translation. We use weight parameter of $\lambda_1 = 1$ in training.

We use two losses for rotation prediction:
Chamfer distance~\cite{fan2017point} $\mathcal{L}_C$ and $L2$ distance $\mathcal{L}_E$. 
Because many parts have symmetric geometry (\eg bars and boards) which results in multiple rotation solutions, we use Chamfer distance as the primary supervising loss to address such pose ambiguity. 
Given the point cloud of part $p_i$, the ground truth rotation $R_i$, and the matched predicted rotation $\hat{R}_{\mathcal{M}(i)}$, 
the Chamfer distance loss is defined as
\begin{equation}
    \mathcal{L}_{C_i} =\frac{1}{d_{pc}}\sum_{x \in \hat{R}_{\mathcal{M}(i)}(p_i)} \min _{y \in R_i(p_i)}\|x-y\|_{2}^{2}+\frac{1}{d_{pc}}\sum_{y \in R_i(p_i)} \min _{x \in \hat{R}_{\mathcal{M}(i)}(p_i)}\|x-y\|_{2}^{2},
\end{equation}\label{eqn:}
where $R_i(p_i)$ and $\hat{R}_{\mathcal{M}(i)}(p_i)$ denote the rotated part point clouds using $R_i$ and $\hat{R}_{\mathcal{M}(i)}$ respectively.
We use $\lambda_2 =20$ for the Chamfer loss. 
Some parts may be not \emph{perfectly} symmetric (\eg one bar that has small but noticeable different geometry at two ends), using Chamfer distance by itself in this case would make the network fall into local minima. 
We encourage the network to correct this situation by penalizing the $L_2$ distance between the matched predicted rotated point cloud and the ground truth rotated point cloud in Euclidean distance.
\begin{equation}
    \mathcal{L}_{E_i} = \frac{1}{d_{pc}}\left\lVert\hat{R}_{\mathcal{M}(i)}(p_i) - R_i(p_i)\right\rVert_F^2,
\end{equation}
where $\left\lVert\cdot\right\rVert_F$ denotes the Frobenius norm, $d_{pc}=1000$ is the number of points per part.
Note that $\mathcal{L}_{E_i}$ on its own is not sufficient in cases when the parts are completely symmetric. Thus, we add the $\mathcal{L}_E$ loss as a regularizing term with a smaller weight of $\lambda_3=1$. We conducted an ablation experiment demonstrating the $\mathcal{L}_E$ loss contributes to correcting rotation for some parts. 

Finally, we compute a shape holistic Chamfer distance as the predicted assembly should be close to the ground truth Chamfer distance.  
\begin{equation}
    \mathcal{L}_{W} =\frac{1}{N\cdot d_{pc}}\sum_{x \in \hat{S}}
     \min _{y \in S}\|x-y\|_{2}^{2}+\frac{1}{N\cdot d_{pc}}\sum_{y \in S} \min _{x \in \hat{S}}\|x-y\|_{2}^{2},
\end{equation}
where $\hat{S} = \cup_{i=1}^{N} (\hat{R}_{\mathcal{M}(i)}(p_i) + \hat{t}_{i})$ denotes the predicted assembled shape point cloud and $S =\cup_{i=1}^{N}(R_i(p_i)+t_i)$ denotes the ground truth shape point cloud.
This loss encourages the holistic shape appearance and the part relationships to be close to the ground-truth.
We use $\lambda_4=1$.

\section{Experiments}
\label{sec:exp}
\vspace{-3mm}

In this section, we set up the testbed for the proposed \emph{single-image-guided 3D part assembly} problem on the PartNet~\cite{Mo:2019:PartNet} dataset. 
To validate the proposed approach, we compare against three baseline methods.
Both qualitative and quantitative results demonstrate the effectiveness of our method. 
\begin{figure}[t]
    \centering
    \vspace{-1mm}
    \includegraphics[width=\linewidth]{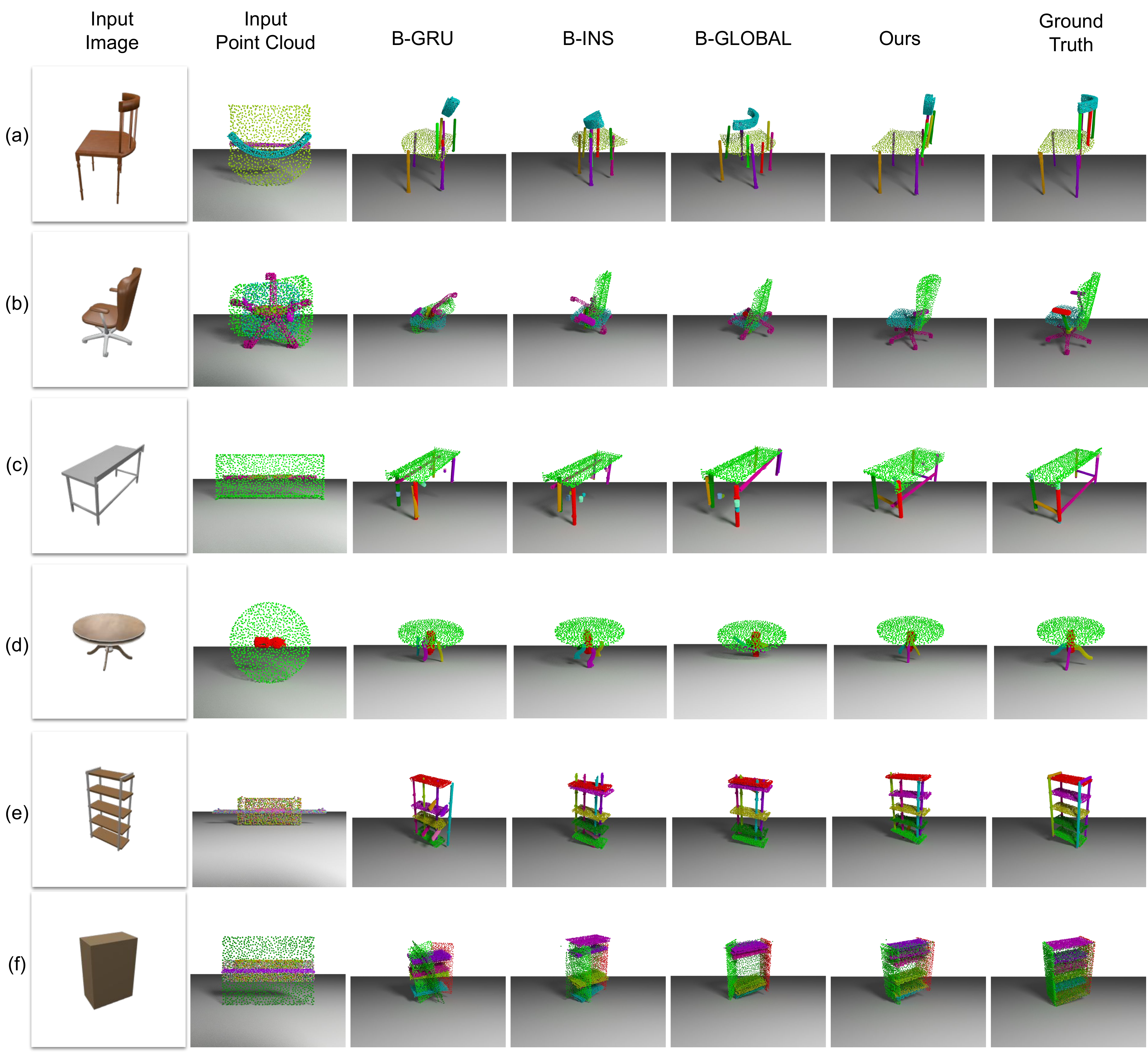}
    \vspace{-4mm}
    \caption{\titlecap{Qualitative Results. } We show six examples of the three furniture categories, each in two different modalities. The upper and lower rows correspond to modality Level-3 and Level mixed respectively.}
    \label{fig:qualitative}
    \vspace{-4mm}
\end{figure}

\subsection{Dataset}
Recently, Mo et. al.~\cite{Mo:2019:PartNet} proposed the PartNet dataset, which is the largest 3D object dataset with fine-grained and hierarchical part annotation.
Every PartNet object is provided with a ground-truth hierarchical part instance-level semantic segmentation, from coarse 
to fine-grained levels
, which provides a good complexity of parts.
In our work, we use the three largest furniture categories that the requires real-world assembly: Chair, Table and Cabinet.
We follow the official PartNet train/validation/test split (roughly $70\%:10\%:20\%$) and filter out the shapes with more than 20 parts. 

For each object category, we create two data modalities: \emph{Level-3} and \emph{Level-mixed}.
The \emph{Level-3} corresponds to the most fine-grained PartNet segmentation.
While we do not assume known part semantics, an algorithm can implicitly learn the semantic priors dealing with the \emph{Level-3} data, which is undesired in our goal of generalizing to real-life assembly settings, as it is unrealistic to assume taht IKEA furnitures also follow the PartNet same semantics. To enforce the network to reason with part geometries, we created an additional category modality, \emph{Level-mixed}, which contains part segmentation at all levels in the PartNet hierarchy. Specifically, for each shape, we traverse every path of the ground-truth part hierarchy and stop at any level randomly.
We have 3736 chairs, 2431 tables, 704 cabinets in \emph{Level-3} and 4664 chairs, 5987 tables, 888 cabinets in \emph{Level-mixed}.

For the input image, we render a set of $224\times224$ images the PartNet models with ShapeNet textures~\cite{ShapeNet}.
We then compute the world-to-camera matrix accordingly and obtain the ground-truth 3D object position in the camera space, which is used for supervising part-instance segmentation supervision.
For the input point cloud, we use Furthest Point Sampling (FPS) to sample $d_{pc}=1000$ points over the each part mesh.
We then normalize them following the descriptions in Sec.~\ref{sec:prod}.
After parts are normalized, we detect geometrically equivalent classes of parts by first filtering out parts comparing dimensions of AABB under a threshold of 0.1. We further process the remaining parts computing all possible pairwise part Chamfer distance normalized by their average diagonal length under a hand-picked threshold of 0.02. 

\subsection{Evaluation Metric}
To evaluate the part assembly performance, we use two metrics: \emph{part accuracy} and \emph{shape Chamfer distance}.
The community of object pose estimation usually uses metrics such as 5-degree-5-cm. 
However, fine-grained part segments usually show abundant pose ambiguity.
For example, a chair leg may be simply a cylinder which has a full rotational and reflective symmetry.
Thus, we introduce the \emph{part accuracy} metric that leverages Chamfer distance between the part point clouds after applying the predicted part pose and the ground truth pose to address such ambiguity. Following previously defined notation in Section 4.3, we define the Part Accuracy Score (PA) as follows and set a threshold of $\tau=0.1$. 

\vspace{-1.3mm}
\begin{equation}
    PA = \frac{1}{N}\sum_{i=1}^{N} \mathbbm{1} \left(\left\lVert( \hat{R}_{\mathcal{M}(i)}(p_i) + \hat{t}_{i}) - ( R_{i}(p_i) + {t}_{i}) \right\rVert_{chamfer} < \tau\right)
\end{equation}

Borrowing the evaluation metric heavily used in the community of 3D object reconstruction, we also measure the \emph{shape Chamfer distance} from the predicted assembled shape to the ground-truth assembly. 
Formally, we define the \emph{shape Chamfer distance} metric $SC$ borrowing notations defined in Section 4.3 as follows.
\vspace{-1.3mm}
\begin{equation}
    SC(S,\hat{S}) = \frac{1}{N\cdot d_{pc}}\sum_{x \in \hat{S}} \min _{y \in S}\|x-y\|_{2}+\frac{1}{N\cdot d_{pc}}\sum_{y \in S} \min _{x \in \hat{S}}\|x-y\|_{2}
\end{equation}

\subsection{Baseline Methods}
We compare our approach to three baseline methods. 
Since there is no direct comparison from previous works that address the exactly same task,
we try to adapt previous works on part-based shape generative modeling~\cite{Wu:2019:pqnet,Sung:2017,Mo:2019:Structurenet,Niu:2018} to our setting and compare with them.
Most of these works require known part semantics and thus perform part-aware shape generation without the input part conditions.
However, in our task, there is no assumption for part semantics or part priors, and thus all methods must explicitly take the part input point clouds as input conditions. We train all three baselines with the same pose loss used in our method defined in Section 4.3.

\mypara{Sequential Pose Proposal (\texttt{B-GRU})}
The first baseline is a sequential model, similar to the method proposed by~\cite{Wu:2019:pqnet,Sung:2017}, instead of sequentially generating parts, we sequentially decode $k$ candidate possible poses for a given part geometry, conditioned on an image. For each input part, if there is $n$ geometrically equivalent parts , where $n \leq k$, we take the first n pose proposal generated using GRU, and conduct Hungarian matching to match with the $n$ ground truth part poses. 

\begin{table}[b]
\caption{Part Accuracy and Assembly Chamfer Distance(CD)} 
\centering
\label{table_1}
{\scriptsize
\begin{tabularx}{\textwidth}{>{\centering}m{1.2cm}|>{\centering}m{2cm}|CCC|CCC}
    \toprule
    \multirow{2}{*}{Modality} & \multirow{2}{*}{Method} & \multicolumn{3}{c|}{Part Accuracy $\uparrow$} & \multicolumn{3}{c}{Assembly CD $\downarrow$} \\
    & & Chair & Table & Cabinet & Chair & Table & Cabinet \\
    \midrule
    \multirow{4}{*}{Level-3} 
    & B-GRU & 0.310 & 0.574 & 0.334  & 0.107 & 0.057 & 0.062\\
    & B-InsOnehot & 0.173 & 0.507 & 0.295 & 0.130 & 0.064 & 0.065\\
    & B-Global & 0.170 & 0.530 & 0.339 & 0.125  & 0.061 & 0.065\\
    & Ours & \textbf{0.454} & \textbf{0.716} & \textbf{0.402} & \textbf{0.067}  & \textbf{0.037} & \textbf{0.050}\\
    \midrule
    \multirow{4}{*}{Mixed} 
    & B-GRU & 0.326 & 0.567 & 0.283 & 0.101 & 0.070 & 0.066\\
    & B-InsOnehot & 0.286 & 0.572 & 0.320 & 0.108 & 0.067 & 0.061\\
    & B-Global & 0.337 & 0.619 &  0.290 & 0.093  & 0.062 & 0.0677\\
    & Ours & \textbf{0.491} & \textbf{0.778} & \textbf{0.483} & \textbf{0.065}  & \textbf{0.037} & \textbf{0.043}\\
    \bottomrule
\end{tabularx}
}
\vspace{-4mm}
\end{table}

\mypara{Instance One-hot Pose Proposal (\texttt{B-InsOneHot})} The second baseline uses MLP to directly infer pose for a given part from its geometry and the input image,
similar to previous works~\cite{Mo:2019:Structurenet,Niu:2018} that output box abstraction for shapes. 
Here, instead of predicting a box for each part, we predict a 6D part pose $(R_j, t_j)$.
We use instance one-hot features to differentiate between the equivalent part point clouds, and conduct Hungarian matching to match with the ground truth part poses regardless of the onehot encoding.

\mypara{Global Feature Model (\texttt{B-Global})} The third baseline is proposed by improving upon the second baseline by adding our the context-aware 3D part feature defined in Section 4.1. Each part pose proposal not only considers the part-specific 3D feature and the 2D image feature, but also a 3D global feature obtained by aggregating the all 3D part feature then max-pool to a global 3D feature containing information of all parts. 
This baseline shares similar ideas to PAGENet~\cite{Li:2020} and CompoNet~\cite{Schor:2019} that also compute global features to assemble each of the generated parts.

\begin{table}[t]
\caption{Visible and Invisible Part Accuracies} 
\centering
\label{table_2}
{\scriptsize
\begin{tabularx}{\textwidth}{>{\centering}m{1.2cm}|>{\centering}m{2cm}|CCC|CCC}
    \toprule
    \multirow{2}{*}{Modality} & \multirow{2}{*}{Method} & \multicolumn{3}{c|}{Part Accuracy (Visible) $\uparrow$} & \multicolumn{3}{c}{Part Accuracy (Invisible) $\uparrow$} \\
    & & Chair & Table & Cabinet & Chair & Table & Cabinet \\
    \midrule
    \multirow{4}{*}{Level-3} 
    & B-GRU & 0.3182 & 0.598 & 0.353  & 0.206 & 0.481 & 0.304\\
    & B-InsOnehot & 0.178 & 0.572 & 0.291 & 0.104 & 0.369 & 0.289\\
    & B-Global & 0.174 & 0.563 & 0.354 & 0.120  & 0.427 & 0.269\\
    & Ours & \textbf{0.471} & \textbf{0.753} & \textbf{0.455} & \textbf{0.270}  & \textbf{0.557} & \textbf{0.358}\\
    \midrule
    \multirow{4}{*}{Mixed} 
    & B-GRU & 0.335 & 0.593 & 0.302 & 0.180 & 0.267 & 0.258\\
    & B-InsOnehot & 0.295 & 0.592 & 0.346 & 0.133 & 0.275 & 0.279\\
    & B-Global & 0.334 & 0.638 &  0.320 & 0.184  & 0.349 & 0.227\\
    & Ours & \textbf{0.505} & \textbf{0.803} & \textbf{0.537} & \textbf{0.262}  & \textbf{0.515} & \textbf{0.360}\\
    \bottomrule
\end{tabularx}
}
\vspace{-4mm}
\end{table}

\subsection{Results and Analysis}
We compare with the three baselines and observe that our method outperforms the baseline methods both qualitatively and quantitatively using the two evaluation metrics, PA and SC. 
We show significant improvement for occluded part pose hallucination as Table~\ref{table_2} demonstrates. Qualitatively, we observe that our method can learn to infer part poses for invisible parts by (1) learning a category prior and (2) leveraging visible parts of the same geometric equivalent class. Our network can reason the stacked placement structure of cabinets as shown in the last row in Fig~\ref{fig:qualitative}. The input image does not reveal the inner structure of the cabinet and our proposed approach learns to vertically distribute the geometrically equivalent boards to fit inside the cabinet walls, similar to the ground truth shape instance. The top row of Fig~\ref{fig:qualitative} demonstrates how our network learns to place the occluded back bar along the visible ones. This could be contributed to our first stage of graph convolution where we leverage visible parts to infer the pose for occluded parts in the same geometrically equivalent class.

Our method demonstrates the most faithful part pose prediction for the shape instance depicted by the input image. As shown in Fig~\ref{fig:qualitative} row (e), our method equally spaces the board parts vertically, which is consistent with the shape structure revealed by the input image. This is likely resulted from our part-instance image segmentation module where we explicitly predict a 2D-3D grounding, whereas the baseline methods lack such components, and we further demonstrate its effectiveness with an ablation experiments.


\begin{figure}[t]
    \centering
    \includegraphics[width=1\linewidth]{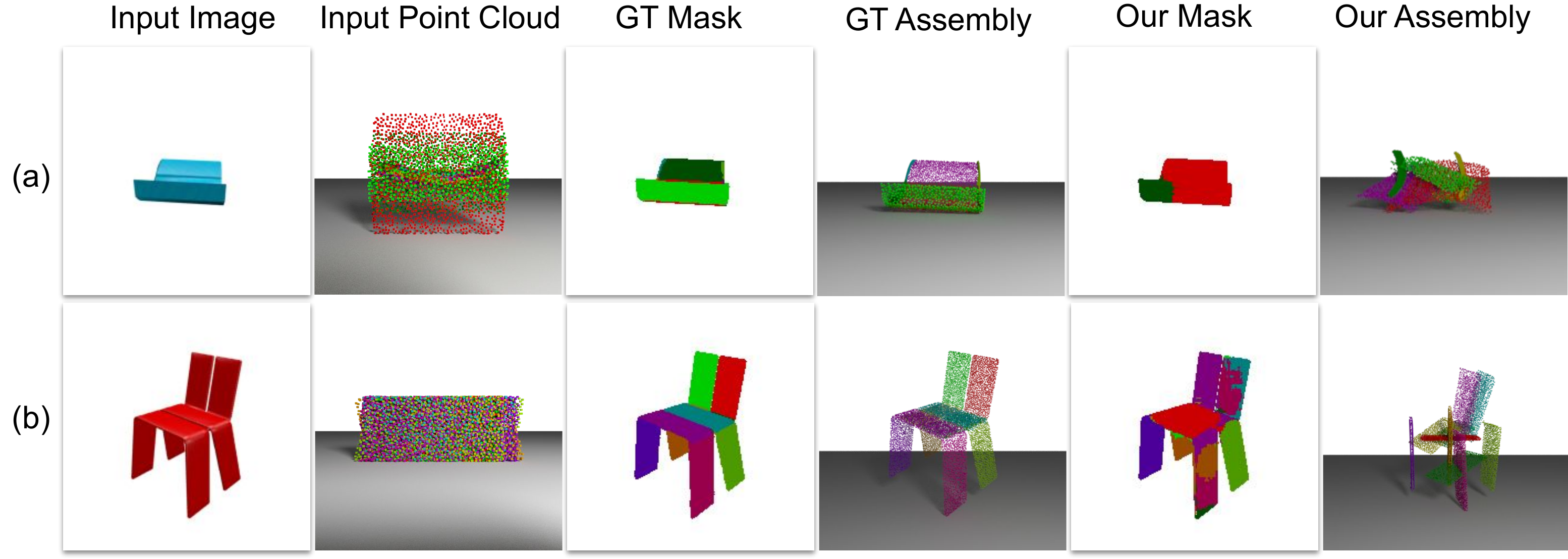}
    \vspace{-6mm}
    \caption{\titlecap{Failure Case}. We show two examples of failure cases. Case (a), the input image is not geometrically informative. Case (b), the chair has only one type of part geometry. }
    \label{fig:failurecase}
    \vspace{-4mm}
\end{figure}

However, our proposed method has its limitations in dealing with unusual image views, exotic shape instance, and shapes composed of only one type of part geometry, which result in noisy mask prediction. The 2D-3D grounding error cascades to later network modules resulting in poor pose predictions. As shown in Fig~\ref{fig:failurecase} row (a), the image view is not very informative of the shape structure, making it difficult to leverage 3D geometric cues to find 2D-3D grounding. Additionally, this chair instance itself is foreign to Chair category. We avoided employing differentiable rendering because it does not help address such failure cases. Fig~\ref{fig:failurecase} row (b) reflects a case where a shape instance is composed of a single modality of part geometry. Geometric affinity of the board parts makes it difficult for the network to come to a determinant answer for the segmentation prediction, resulting in a sub-optimal part pose prediction. These obstacles arise from the task itself that all baselines also suffer from the same difficulties. 

\mypara{Ablation Experiments}
We conduct several ablation experiments on our proposed method and losses trained on PartNet Chair Level-3. Table~\ref{tab_ablation} in Appendix demonstrates the effectiveness of each ablated component. The part-instance image segmentation module plays the most important role in our pipeline. Removing it results in the most significant performance decrease.

\section{Conclusion and Future Works}
\label{sec:conclusion}
We formulated a novel problem of \emph{single-image-guided 3D part assembly} and proposed a neural-net-based pipeline for the task that leverages information from both 2D grounding and 3D geometric reasoning.
We established a test bed on the PartNet dataset.
Quantitative evaluation demonstrates that the proposed method achieves a significant improvement upon three baseline methods. 
For the future works, one can study how to leverage multiple images or 3D partial scans as inputs to achieve better results.
We also do not explicitly consider the connecting junctions between parts (\eg pegs and holes) in our framework, which are strong constraints for real-world robotic assembly.

\section{Acknowledgements}
We thank the Vannevar Bush Faculty Fellowship and the grants from the Samsung GRO program and the SAIL Toyota Research Center for supporting the authors with their research, but this article reflects only the opinions and conclustions of its authors. We also thank Autodesk and Adobe for the research gifts provided. 

\section{Appendix}
\label{sec:appendix}
\noindent This document provides supplementary materials accompanying the main paper, including
\begin{itemize}
    \item Ablation Experiments
    \item Discussion of failure cases and future works;
    \item More Architecture Details;
    \item More Qualitative Examples.
\end{itemize}
\subsection*{A. Ablation Experiments}
\begin{center}
\begin{table}
    \vspace{-0.5\baselineskip}
    \vspace{-4mm}
    \centering
    \caption{Ablation Experiment Results} 
    \label{tab_ablation}
    {\footnotesize
    \begin{tabularx}{\textwidth}{>{\centering}m{3cm}|>{\centering}m{1.8cm}CC|C}
        \toprule
        \begin{tabular}[c]{@{}c@{}}{Ablated}\\{Module}\end{tabular} & \begin{tabular}[c]{@{}c@{}}{}\\Total\end{tabular}   & 
        \begin{tabular}[c]{@{}c@{}}Part Accuracies $\uparrow$\\Visible\end{tabular}
        &
        \begin{tabular}[c]{@{}c@{}}{ }\\Invisible\end{tabular}
        &
    \begin{tabular}[c]{@{}c@{}} Assembly CD $\downarrow$\end{tabular}\\
    \midrule

  w/o L2 Rotation loss & 0.426 & 0.445 & 0.207  & 0.070\\
    w/o Segmentation & 0.363 & 0.378 & 0.164& 0.084 \\
   w/o Graph Conv 1, 2& 0.403 & 0.423 & 0.178 & 0.073\\
  w/o Graph Conv 2 & 0.434  & 0.456 & 0.239 & 0.073 \\
   w/o Image Feature & 0.403  & 0.419 & 0.208 & 0.077\\
   w/o Global Feature &0.418 & 0.437 & 0.202 & 0.072\\
   \midrule
   Ours - Full & \textbf{0.454} & \textbf{0.470} & \textbf{0.270} & \textbf{0.067}\\     \bottomrule
   \end{tabularx}
}
\vspace{-8mm}
\end{table}

\end{center}
\subsection*{B. More Failure Cases and Discussion}

\begin{figure}[t]
    \centering
    \includegraphics[width=1\linewidth]{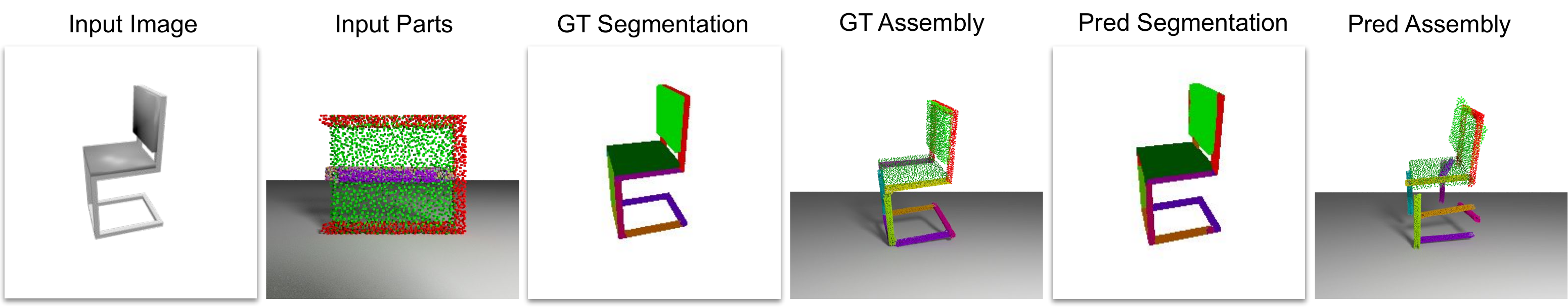}
    \vspace{-6mm}
    \caption{\titlecap{Failure Case} This figure shows that our proposed method does not well handle disconnected parts, and needs to leverage more geometric reasoning.}
    \label{fig:failurecaseb3}
    \vspace{-4mm}
\end{figure}

\mypara{Disconnected Parts} We notice that our prediction on very fine-grained instances sometimes results in unconnected parts. The assembly setting requires the physical constraint that each part must be in contact with another part. However, the implicit soft constraint enforced using the second stage graph graph convolution is not sufficient enough for this task. Ideally, the translation and rotation predicted for each part is only valid if they can transform the part to be in contact at the joints between relevant parts. For example, in Figure~\ref{fig:failurecaseb3} we can see that the back of the chair base bars does not connect. We plan to address this problem in future works by explicitly enforcing contact between parts in a range of contact neighborhood. 

\mypara{Geometric Reasoning} Additionally, though our current proposed method makes many design choices geared for geometric reasoning between fitting of parts, however, we still see some cases that the fitting between parts is not yet perfect. For example, in Figure~\ref{fig:failurecaseb3}, We can see that the back pad does not fit perfectly into the back frame bar. This problem need to be addressed in future work where the method design should discover some pairwise or triplet-level geometric properties that allow fitting between parts. 

\newpage
\subsection*{C. Architecture Details}
\begin{center}
  
  \begin{table}[!h]
    \centering
    \footnotesize
    \vspace{-0.5mm}
    \caption{Part-instance Segmentation Architecture.}
    
    \begin{tabular}{c|l}
        \noalign{\hrule height 1pt}
        layer & configuration \\
        \noalign{\hrule height 1pt}
        \multicolumn{2}{c}{UNet Encoding} \\
        \noalign{\hrule height 1pt}
        \centering
        1 & Conv2D (3, 32, 3, 1, 1), ReLU, BN, \\ &Conv2D (32, 32, 3, 1, 1), ReLU, BN, \\
        \hline
        2 & Conv2D (32, 64, 3, 1, 1), ReLU, BN, \\& Conv2D (64, 64, 3, 1, 1), ReLU, BN,  \\
        \hline
        3 & Conv2D (64, 128, 3, 1, 1), ReLU, BN, \\& Conv2D (128, 128, 3, 1, 1), ReLU, BN,  \\
        \hline
        4 & Conv2D (128, 256, 3, 1, 1), ReLU, BN, \\& Conv2D (256, 256, 3, 1, 1), ReLU, BN,  \\
        \hline
        5 & Conv2D (256, 512, 3, 1, 1), ReLU, BN, \\&  Conv2D (512, 512, 3, 1, 1), ReLU, BN,  \\
        
        \noalign{\hrule height 1pt}
        \multicolumn{2}{c}{UNet Decoding} \\
        \noalign{\hrule height 1pt}
        \centering
        1 & ConvTranspose2D(1301, 256, 2, 2) \\
        \hline
        2 & ConvTranspose2D(256, 128, 2, 2)\\
        \hline
        3 & ConvTranspose2D(128, 64 , 2, 2)\\
        \hline
        4 & ConvTranspose2D(64, 32, 2, 2)\\
        \hline
        5 & ConvTranspose2D(32, 1, 1, 1)\\
        \hline
        \noalign{\hrule height 1pt}     
        \multicolumn{2}{c}{PointNet} \\
        \noalign{\hrule height 1pt}
        1 & Conv1D (3, 64, 1, 1), BN, ReLU\\
        \hline
        2 & Conv1D (64, 64, 1, 1), BN, ReLU\\
        \hline
        3 & Conv1D (64, 64, 1, 1), BN, ReLU\\
        \hline
        4 & Conv1D (64, 128, 1, 1), BN, ReLU\\
        \hline
        5 & Conv1D (128, 512, 1, 1), BN, ReLU\\
        \noalign{\hrule height 1pt}     
        \multicolumn{2}{c}{SLP1} \\
        \noalign{\hrule height 1pt}
        1 & FC (512, 256), ReLU, MaxPool1D\\
        \noalign{\hrule height 1pt}
        \multicolumn{2}{c}{SLP2} \\
        \noalign{\hrule height 1pt}
        1 & FC (256, 256), ReLU \\
        \noalign{\hrule height 1pt} 
    \end{tabular}
     
        \label{tab:arch_mask}
\end{table}
  
  \begin{table}[!h]
    \centering
    \vspace{-2mm}
    \caption{Pose Prediction Architecture.}
    \footnotesize
    \begin{tabular}{c|l}
        \noalign{\hrule height 1pt}
        layer & configuration \\
        \noalign{\hrule height 1pt}
        \multicolumn{2}{c}{SLP 3} \\
        \noalign{\hrule height 1pt}
        \centering
        1 & FC(1301, 256), ReLU \\
        \hline
        \noalign{\hrule height 1pt}
        \multicolumn{2}{c}{Pose Decoder 2} \\
        \noalign{\hrule height 1pt}
        \centering
        1 & FC (1301, 256), ReLU \\
        \hline
        2 & FC(256, 3) \\
        \hline
        3 & FC(256, 4) \\
        
        \noalign{\hrule height 1pt}
        \multicolumn{2}{c}{SLP 4} \\
        \noalign{\hrule height 1pt}
        \centering
        1 & FC(1031, 256), ReLU \\
        \noalign{\hrule height 1pt}
        \multicolumn{2}{c}{Pose Decoder 2} \\
        \noalign{\hrule height 1pt}
        \centering
        1 & FC (1031, 256), ReLU \\
        \hline
        2 & FC(256, 3) \\
        \hline
        3 & FC(256, 4) \\
        \noalign{\hrule height 1pt}
    \end{tabular}
     
        \label{tab:arch_mask}
\end{table}
\end{center}
\subsection*{D. More Qualitative Results}
\begin{figure}[t]
    \centering
    \includegraphics[width=0.97\linewidth]{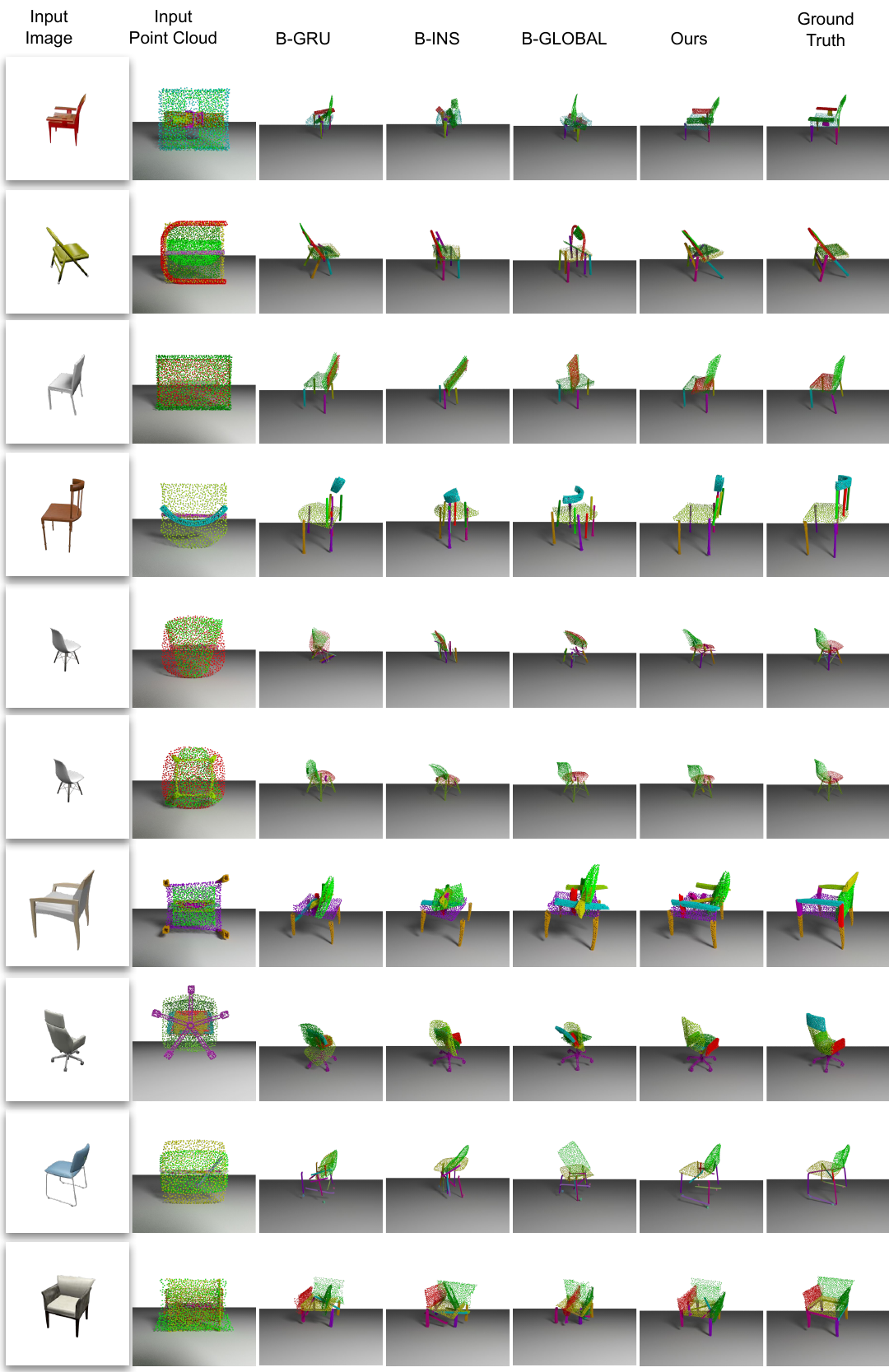}
    \caption{\titlecap{Qualitative Results for the Chair Category.} The top 5 rows show the results of Chair Level-3, and the bottom 5 rows contains the results of Chair Level-mixed.}
    \label{fig:chair}
    \vspace{-3mm}
\end{figure}

\begin{figure}[t]
    \centering
    \includegraphics[width=0.97\linewidth]{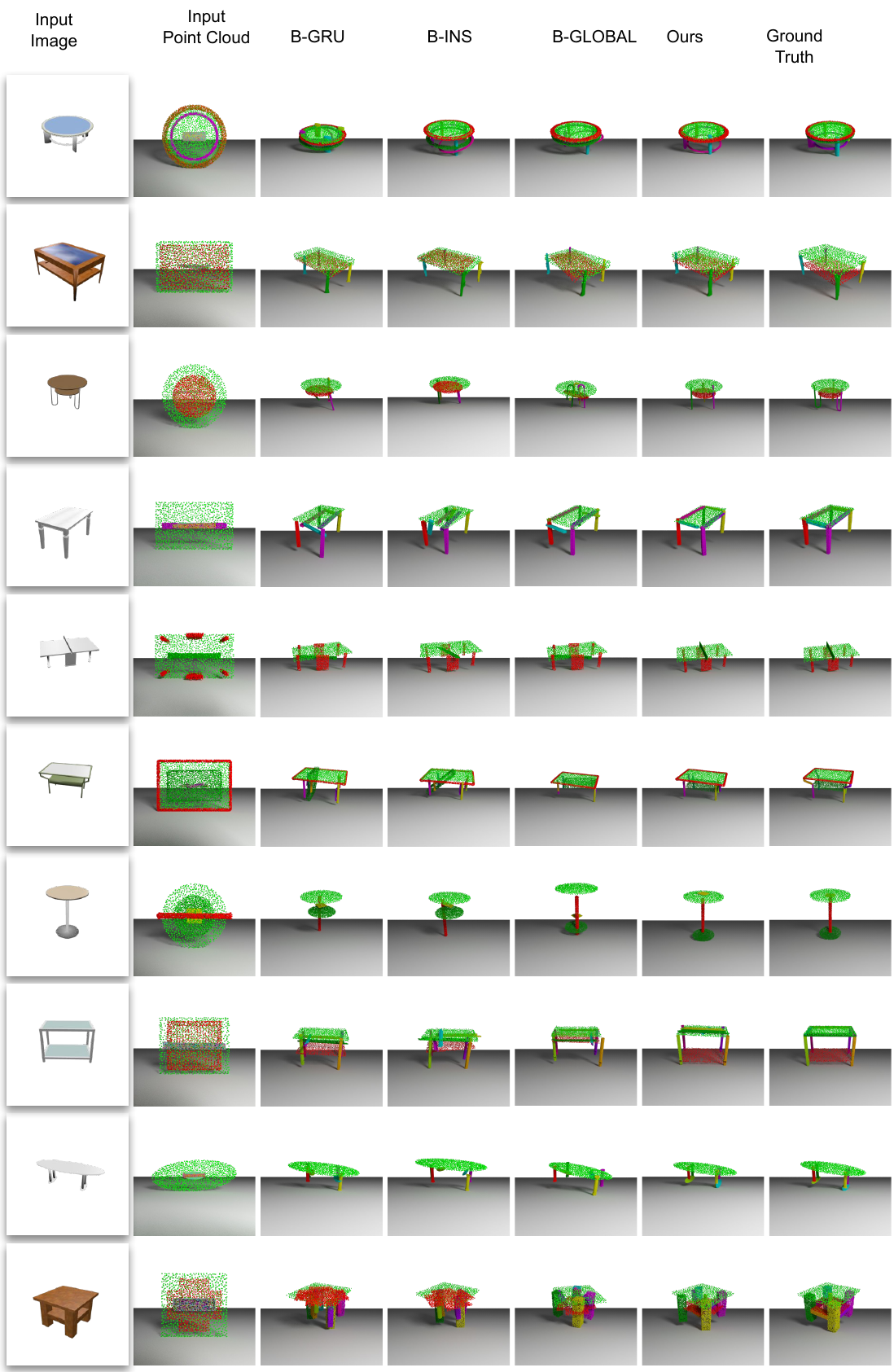}
    \caption{\titlecap{Qualitative Results for the Table Category.} The top 5 rows show the results of Table Level-3, and the bottom 5 rows contains the results of Table Level-mixed.}
    \label{fig:table}
    \vspace{-3mm}
\end{figure}

\begin{figure}[t]
    \centering
    \includegraphics[width=0.95\linewidth]{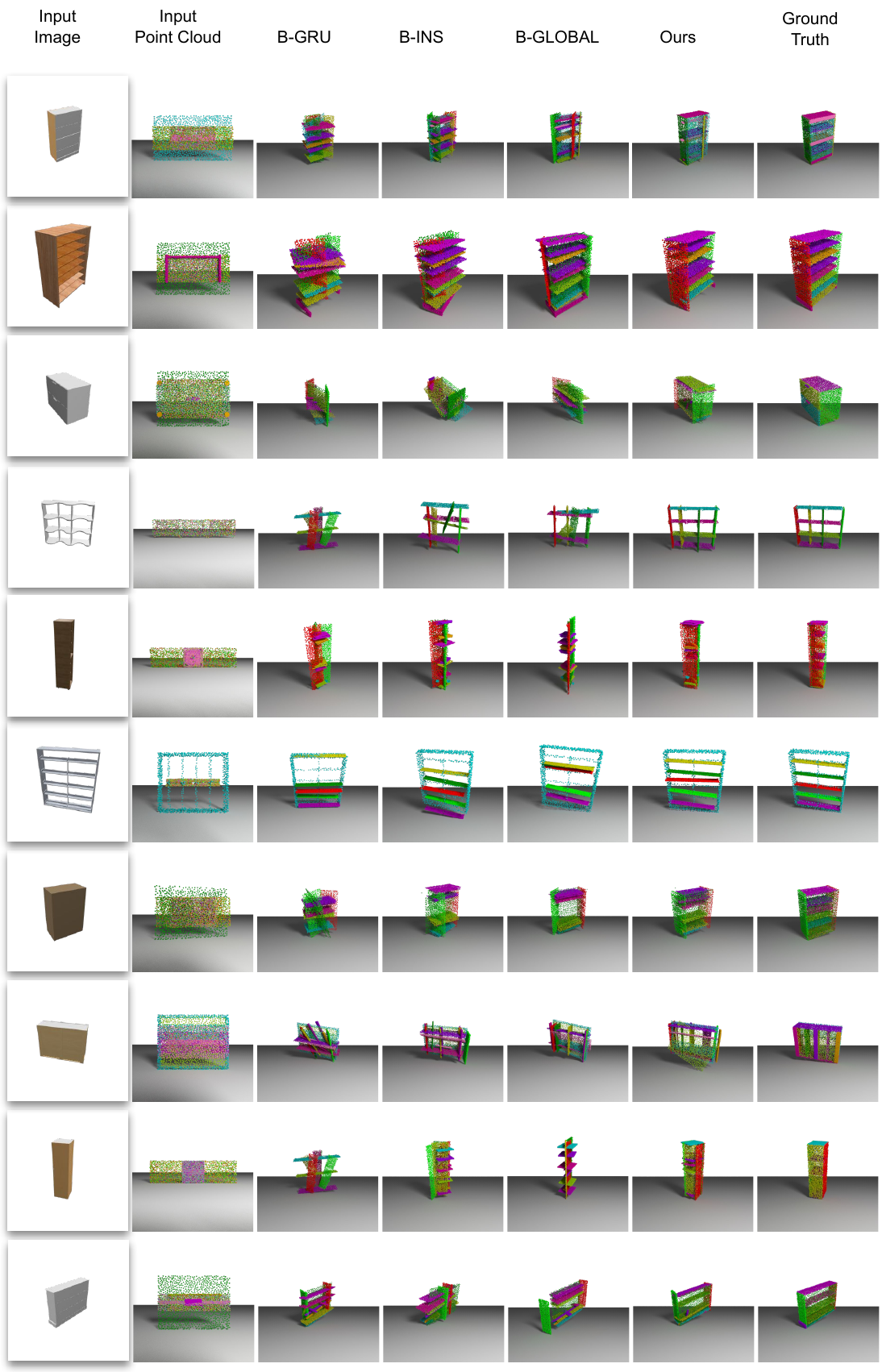}
    \caption{\titlecap{Qualitative Results for the Cabinet Category.} The top 5 rows show the results of Cabinet Level-3, and the bottom 5 rows contains the results of Cabinet Level-mixed.}
    \label{fig:cabinet}
    \vspace{-3mm}
\end{figure}

%
%

\clearpage
\newpage

\bibliographystyle{splncs04}
\bibliography{main}
\end{document}